%% file: iclr2026_conference.tex
\documentclass{article} 
\usepackage{iclr2026_conference,times}
\input{math_commands.tex}

\usepackage{booktabs}
\usepackage{hyperref}
\usepackage{url}
\usepackage[most]{tcolorbox}
\usepackage{multirow}
\usepackage{lineno}
\usepackage{pgf}
\usepackage{pgfplots}
\usepackage{booktabs}
\usepackage{scalerel}
\usepackage{float}
\usepackage[table,dvipsnames]{xcolor}
\usepackage{subcaption}
\usepackage{amssymb}
\usepackage{enumitem}
\usepackage{algorithm}          
\usepackage{algpseudocode}

\newtcolorbox{findingbox}{
    colframe=black,             
    colback=white!,      
    boxrule=0.5pt,
    arc=2pt,
    left=5pt,
    right=5pt,
    top=3pt,
    bottom=3pt,
    boxsep=0pt,
    before skip=8pt,
    after skip=8pt,
    fontupper=\itshape,
    title=Finding 1,
    coltitle=black,             
    attach boxed title to top left={xshift=5pt,yshift=-3pt},
    boxed title style={colframe=black, colback=white!95!gray}
}

\title{\textbf{Co}ntext \textbf{T}okens are \textbf{A}nchors: Understanding the Repetition Curse in dMLLMs from an Information Flow Perspective}

\author{
\textbf{Qiyan Zhao}$^{1,2,5*}$
\textbf{, Xiaofeng Zhang}$^{1*\ddagger}$ 
\textbf{, Shuochen Chang}$^{1}$
\textbf{, Qianyu Chen}$^{3}$
\textbf{, Xuhang Chen}$^{1}$ \\
\textbf{Xiaosong Yuan}$^{1}$ 
\textbf{, Luoqi Liu}$^{4}$ 
\textbf{, Da-Han Wang}$^{2, 5}$
\textbf{, Jiajun Zhang}$^{2}$ 
\textbf{, Xu-Yao Zhang}$^{2}$ 
 \\
\\[-0.25em]
$^{1}$SJTU \quad
$^{2}$CASIA \quad
$^{3}$NTU \quad
$^{4}$Meitu (China) Limited \quad
$^{5}$XMUT \\
$^*$Equal contribution \quad 
$^\ddagger$Corresponding author  \\
(zhaoqiyan2022@163.com, SemiZxf@163.com)\\ 
}

\iclrfinalcopy 

\begin{document}

\maketitle

\begin{abstract}

Recent diffusion-based Multimodal Large Language Models (dMLLMs) suffer from high inference latency and therefore rely on caching techniques to accelerate decoding. However, the application of cache mechanisms often introduces undesirable repetitive text generation, a phenomenon we term the \textbf{Repeat Curse}. To better investigate underlying mechanism behind this issue, we analyze repetition generation through the lens of information flow. Our work reveals three key findings: (1) context tokens aggregate semantic information as anchors and guide the final predictions; (2) as information propagates across layers, the entropy of context tokens converges in deeper layers, reflecting the model’s growing prediction certainty; (3) Repetition is typically linked to disruptions in the information flow of context tokens and to the inability of their entropy to converge in deeper layers. Based on these insights, we present \textbf{CoTA}, a plug-and-play method for mitigating repetition. CoTA enhances the attention of context tokens to preserve intrinsic information flow patterns, while introducing a penalty term to the confidence score during decoding to avoid outputs driven by uncertain context tokens. With extensive experiments, CoTA demonstrates significant effectiveness in alleviating repetition and achieves consistent performance improvements on general tasks. Code is available at \url{https://github.com/ErikZ719/CoTA}

\end{abstract}

\section{Introduction}

Recent advances in diffusion-based large language models (dLLMs)~\cite{llada, llada1.5, dream2025, gong2025scaling} have demonstrated impressive reasoning and parallel decoding capabilities. Unlike autoregressive large language models~\cite{llama, qwen2.5, vicuna}, which generate text by sequentially predicting the next token, dLLMs produce tokens in parallel by framing response generation as an iterative denoising process over a fully masked discrete token sequence. Building on this foundation, diffusion-based Multimodal Large Language Models (dMLLMs)~\cite{lladav, mmada, lavida} have emerged as powerful multimodal systems that integrate vision instruction tuning with dLLMs, and have achieved performance comparable to leading autoregressive architectures across multiple benchmarks~\cite{MMBench, mmstar, mme}. 

\begin{figure}[t]
\centerline{\includegraphics[width=13cm]{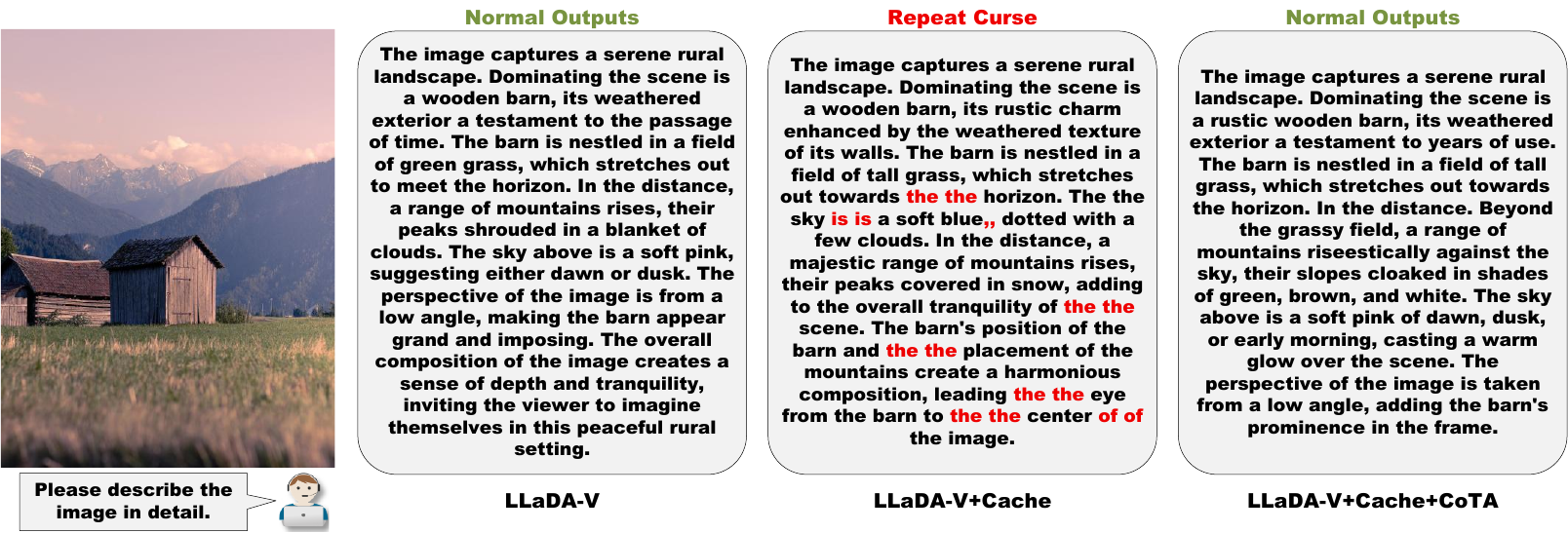}}
\caption{\textbf{Motivation. } When cache is applied to accelerate dMLLMs, the generated responses often exhibit excessive token repetition—a phenomenon we term the Repeat Curse.}
\label{motivation}
\end{figure}

\begin{figure}[t]
\centerline{\includegraphics[width=13cm]{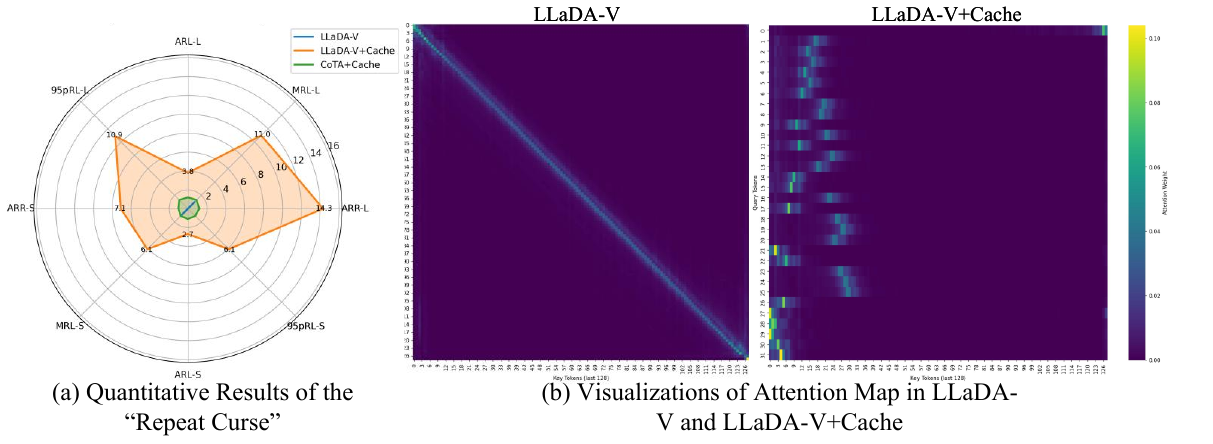}}
\caption{(a) presents the quantitative results of the “Repeat Curse”. L and S indicate evaluations on long-text responses (512 tokens) and short-text responses (64 tokens). The evaluation metrics ARR, MRL, ARL, and 95pRL are introduced in Section~\ref{metric} and Appendix~\ref{appendix2}. (b) visualizes the attention maps of LLaDA-V and LLaDA-V+Cache.}
\label{Repeat}
\end{figure}

Current efforts to accelerate dMLLMs~\cite{dLLMCache, Wu2025FastdLLMTA} primarily exploit customized caching strategies for both prefix and suffix tokens. While these approaches effectively reduce inference latency, our experiments demonstrate that they often introduce a severe side effect: the generated text exhibits substantial repetition, as shown in Figure~\ref{motivation}. This repetition significantly reduces the performance and readability of the model outputs. We refer to this phenomenon as \textbf{“Repeat Curse”}. We propose four complementary metrics to quantitatively evaluate the “Repeat Curse” phenomenon. As illustrated in Figure~\ref{Repeat}.a, our empirical analysis shows that this issue consistently emerges in dMLLMs when cache techniques are employed.

However, the inherent black-box nature of dMLLMs presents a major obstacle to uncovering the internal mechanisms responsible for Repeat Curse. Recent advances in information flow analysis~\cite{Yu2024UnveilingAH, Wang2023LabelWA, fastv} have introduced an interpretable approach for understanding the relationship between model outputs and internal mechanisms, which has motivated the development of numerous methods for mitigating abnormal output patterns (e.g., hallucination~\cite{Tang2025SeeingFA} and degeneration~\cite{Yona2025InterpretingTR}). Inspired by these successes, this work conducts an in-depth analysis from the perspective of information flow to reveal the connection between the internal mechanisms of dMLLMs and the “Repeat Curse.”

By visualizing the attention interaction patterns among tokens (as shown in Figure~\ref{context}), we observe that in dMLLMs, the bidirectional attention mechanism enables context tokens adjacent to the query to act as anchors that progressively aggregate semantic information across layers, \textbf{causing attention to gradually concentrate on these context tokens}. We further introduce information entropy to analyze the impact of context tokens on decoding, as illustrated in \textcolor{red}{Figure~\ref{duibi} a}. We find that \textbf{the entropy of context tokens converges in deeper layers}, reflecting that as information is progressively aggregated across layers, the model’s predictive certainty consistently increases. When repetition arises after applying the cache, the model exhibits random attention allocation \textcolor{red}{(Figure~\ref{Repeat}.b)}. In addition, context tokens corresponding to repeated outputs sustain abnormally high entropy in deeper layers \textcolor{red}{(Figure~\ref{duibi}.b)}. Together, these findings suggest that caching disrupts the inherent mechanisms of dMLLMs, thereby triggering the Repeat Curse. We attribute this phenomenon to two key factors: 1. The introduction of caching disrupts the attention distribution and the inherent information flow patterns of context tokens; 2. Some context tokens whose entropy fails to converge in deeper layers induce the model to generate uncertain tokens, which are often accompanied by repetition.



Building on these insights, we propose \textbf{CoTA}\footnote{The name CoTA is derived from our key finding: \textbf{CO}ntext \textbf{T}okens are \textbf{A}nchors.}, a plug-and-play approach that addresses the abnormal patterns underlying the Repeat Curse and mitigates repetition. CoTA is built on two key components:
(1) Context-token Attention Enhancement (CTAE): a distance-aware attention intervention that strengthens attention to context tokens, thereby preserving the intended information flow during token interactions;
(2) Context-token Entropy-guided Voting (CTEV): a mechanism that leverages the aggregated deep-layer entropy of context tokens as a penalty term in the confidence score, discouraging the model from generating uncertain and repetitive outputs.
CoTA can be seamlessly integrated with baseline dMLLMs and existing caching strategies in a training-free manner, while incurring only modest computational overhead. Experimental results show that CoTA is highly effective in mitigating repetition, reducing the adjacent repetition rate by up to 92\%. Furthermore, across several multimodal benchmarks, CoTA consistently surpasses the baseline, demonstrating substantial improvements in overall robustness and generalization.


Our contributions are three-fold. 

\begin{itemize}[leftmargin=*]
\item First, we identify the Repeat Curse phenomenon that emerges when caching is applied in dMLLMs and uncover its underlying causes through information flow analysis.
\item Second, we introduce CoTA, a plug-and-play approach specifically designed to alleviate repetition.
\item Third, we validate the effectiveness of CoTA through extensive experiments, demonstrating consistent performance improvements across multiple general multimodal tasks.
\end{itemize}

\section{Related Work}

\noindent
\textbf{Diffusion-based Multimodal Large Language Models.} The latest diffusion-based large language models (dLLMs)~\cite{llada, llada1.5, dream2025, gong2025scaling} have been successfully scaled to 8B parameters, achieving performance comparable to state-of-the-art autoregressive large language models~\cite{llama3, deepseek, qwen2.5}. By combining visual instruction tuning~\cite{llava1.5} with dLLMs, ~\cite{lladav, mmada, lavida} successfully develops diffusion-based multimodal large language models.

\noindent
\textbf{Information Flow.} Recent studies have underscored the importance of information flow as an intuitive means of representing the internal mechanisms of black-box models. Common approaches for analyzing information flow mainly include saliency scores~\cite{Yu2024UnveilingAH, Wang2023LabelWA}, attention maps~\cite{Xiao2023EfficientSL, OPERA}, Grad-CAM~\cite{Zhang2024FromRT}, and massive values~\cite{Jin2025MassiveVI}, among others. Prior studies~\cite{Yu2024UnveilingAH, fastv, tame} have demonstrated the existence of certain anchor tokens in autoregressive models, which aggregate information and play a crucial role in cross-layer information flow. \textcolor{red}{Additionally, ADLM \cite{ADLM} discusses the role of anchors in semantic guidance within diffusion language models.} These findings inspire our exploration of information flow in dMLLMs.

\section{Motivation and Analysis}\label{moti}

In this section, we begin with a brief overview of baseline dMLLMs and the cache mechanism\footnote{This paper adopts LLaDA-V~\cite{lladav} and dLLM-Cache~\cite{dLLMCache} as the baseline dMLLMs and cache method.}. We then investigate the ‘Repeat Curse’ through both quantitative and qualitative analyses. Finally, we compare the information flow in dMLLMs with and without cache, shedding light on the underlying cause of the Repeat Curse.

\subsection{Preliminary}

\noindent
\textbf{Diffusion-based Multimodal Large Language Models (dMLLMs).} Typically, a dMLLM $\mathcal{F}$ consists of three main components: a pretrained vision encoder $\mathcal{F}_v$, a dLLM $\mathcal{F}_t$, and a projector $f$ that maps visual features into the text embedding space. Given an image input $I_v$ and an instruction prompt $I_p$ (e.g., \textit{``Please describe the image in detail''}), the model converts them into tokens and concatenates them into a multimodal sequence: $\{\mathcal{S}_{v}, \mathcal{S}_{t}\}$, where
$ \mathcal{S}_{v} = f(\mathcal{F}_v(I_v)) =\{ w_{i}\}^V_{i=1}$ and $ \mathcal{S}_{t} = \mathcal{F}_t(I_p)=\{ w_{i}\}^T_{i=1}$ represent visual and instruction tokens of lengths $V$ and $T$, respectively. Subsequently, a fully masked token sequence $\mathcal{S}_{m} = \{ w_{i}\}^M_{i=1}$ of lengths $M$ is initialized as the response sequence and concatenated with $\{\mathcal{S}_{v}, \mathcal{S}_{t}\}$ to form the final input sequence $\mathcal{S} =\{\mathcal{S}_{v}, \mathcal{S}_{t}, \mathcal{S}_{m}\}$. In such sequence, different tokens share the same dimension.

\begin{figure}[t]
\centerline{\includegraphics[width=13cm]{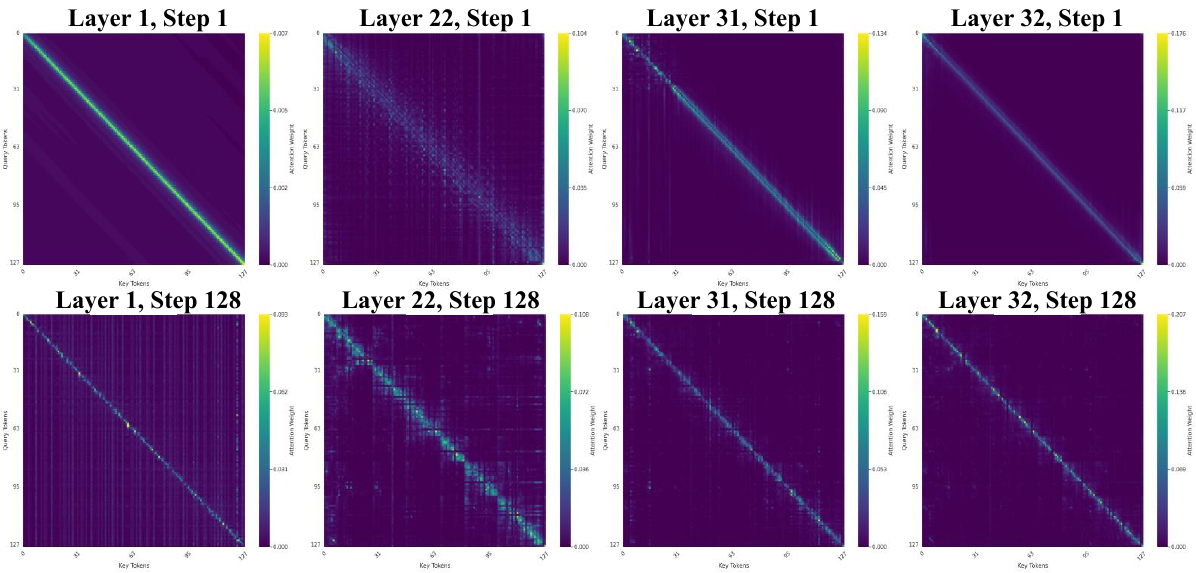}}
\caption{\textbf{\textcolor{red}{Attention Maps Visualization of LLaDA-V}}. Based on LLaDA-V 8B with a generation length of 128 and 32 decoding steps, we visualize the attention matrices corresponding to the masked (response) tokens, with the x-axis denoting key tokens and the y-axis denoting query tokens. Brighter colors indicate higher attention values. Context tokens act as anchors to aggregate information across layers and absorb attention.}
\label{context}
\end{figure}

\noindent
\textbf{Inference and Sampling in dMLLMs.} 
Let $\mathcal{S}^{t}=\{\mathcal{S}_{v}^{t}, \mathcal{S}_{t}^{t}, \mathcal{S}_{m}^{t}\}$ denote the input sequence state at decoding step $t$. 
Starting from $\mathcal{S}^{0}$, where the response sequence $\mathcal{S}_{m}$ is fully masked, the dMLLM performs an iterative unmasking process over $T$ discrete steps to generate the final text response. 
At each step $t$, the model computes a probability distribution $p_{\theta}(\mathcal{S}^{t} \mid \mathcal{S}^{t-1})$ for every masked token. 
From these distributions, the most likely token predictions $\hat{\mathcal{S}}^{t}_{(i)}$ and their corresponding confidence scores $c_{(i)}$ are determined:
\begin{equation}
\hat{\mathcal{S}}^{t}_{(i)} = \arg\max_{v \in V} p_{\theta}(\mathcal{S}^{t}_{(i)} = v \mid \mathcal{S}^{t-1}) 
\quad \text{and} \quad 
c_{(i)} = p_{\theta}(\mathcal{S}^{t}_{(i)} = \hat{\mathcal{S}}^{t}_{(i)} \mid \mathcal{S}^{t-1}), \quad i\in M_{t-1},
\label{ci}
\end{equation}
where $M_{t-1}$ represents the index set of masked token positions at step $(t-1)$ and $V$ is the vocabulary. 
Finally, the model selects the $k$ positions in $M_{t-1}$ with the highest confidence scores $c_{(i)}$ and obtains the set of indices to update $U_{t}$. 
The $k$ selected tokens are unmasked at this step, producing the updated sequence:

\begin{equation}
\mathcal{S}^{t}_{(i)} =
\begin{cases}
\hat{\mathcal{S}}^{t}_{(i)}, & \text{if } i \in U_{t}, \\
\mathcal{S}^{t-1}_{(i)}, & \text{otherwise}.
\end{cases}
\end{equation}

The indices of the masked tokens at step $t$ are updated by $M_{t} = M_{t-1} - U_{t}$.

\noindent
\textbf{Cache Mechanism for dMLLMs.} During the iterative unmasking process of dMLLMs, attention needs to be computed over all tokens in the sequence $\mathcal{S}$ at each step, which leads to significant inference latency. Since dMLLMs adopt a bidirectional attention mechanism, the conventional KV-cache technique~\cite{xiao2024streamingllm} is not applicable. To address this, several caching methods~\cite{dLLMCache, Wu2025FastdLLMTA} leverage a similar observation: prefix tokens $\{\mathcal{S}_{v}, \mathcal{S}_{t}\}$ and parts of the suffix tokens $\mathcal{S}_{m}$ exhibit minimal changes in their attention values during inference, which enables the design of token state caching and reuse. For example, in~\cite{dLLMCache}, the core caching mechanism is defined as follows:
\begin{equation}
\begin{cases}
\text{Recompute}(\{\mathcal{S}_{v}^t, \mathcal{S}_{t}^t\}), & \text{if } t \bmod \tfrac{T}{\mathcal{E}_p} = 0, \\[6pt]
\text{Recompute}(\mathcal{S}_{m}^t), & \text{if } t \bmod \tfrac{T}{\mathcal{E}_s} = 0, \\[6pt]
\text{Recompute}(\mathcal{S}_{(i)}^t), & \text{if } \mathcal{S}_{(i)}^t \in \operatorname*{arg\,min}_{\alpha}\!\Big( \operatorname{Sim}\!\big(\mathcal{S}_{m}^t, \mathcal{S}_{m}^{t-1}\big) \Big). \\[6pt]
\end{cases}
\label{4}
\end{equation}
Here, $\operatorname{Recompute}()$ denotes the recomputation of attention. $\mathcal{E}_p$ and $\mathcal{E}_s$ indicate the predefined update periods for prefix and suffix tokens, respectively. $\operatorname*{arg\,min}_{\alpha}$ selects the bottom $\alpha$ proportion of elements from the sequence, and $\operatorname{Sim}()$ refers to the cosine similarity function.\footnote{We adopt dLLM-Cache~\cite{dLLMCache} as the baseline cache method, keeping all relevant settings strictly consistent to ensure experimental fairness. Hyperparameters are fixed as $\alpha = 25\%$, $\mathcal{E}_p = 25$, and $\mathcal{E}_s = 7$.}

\subsection{Repeat Curse}\label{metric}

\noindent
\textbf{Repetition Quantitative Metrics.} As illustrated in Figure~\ref{motivation}, we define the redundant repetition of tokens in model responses as the Repeat Curse. To quantitatively evaluate this phenomenon, we introduce the Adjacent Repetition Rate (ARR), which measures the proportion of repeated tokens within a response sequence $\{ y_{i}\}^M_{i=1}$ of lengths $M$:
\begin{equation}
\text{ARR} = \frac{1}{M-1} \sum_{i=1}^{M} \mathbf{1}\!\left( y_{i} = y_{i-1} \right),
\end{equation}
where $\mathbf{1}()$ is an indicator function that takes the value 1 when the condition is satisfied and 0 otherwise. In addition, we introduce sample repetition rate (SRR), maximum repetition length (MRL), average repetition length (ARL), and 95th-percentile repetition length (95pRL) to evaluate the severity of the Repeat Curse. (The detailed computation procedures are presented in Appendix~\ref{appendix2}.)

\noindent
\textbf{Quantitative Experimental Results.} Figure~\ref{Repeat}(a) summarizes the quantitative results of the Repeat Curse under both long- and short-response settings. The experiments are performed on the image captioning task with 500 randomly sampled MSCOCO images~\cite{Lin2014MicrosoftCC}. We observe that the baseline model suffers from severe token repetition when the cache is applied.

\begin{figure}[t]
\centerline{\includegraphics[width=14cm]{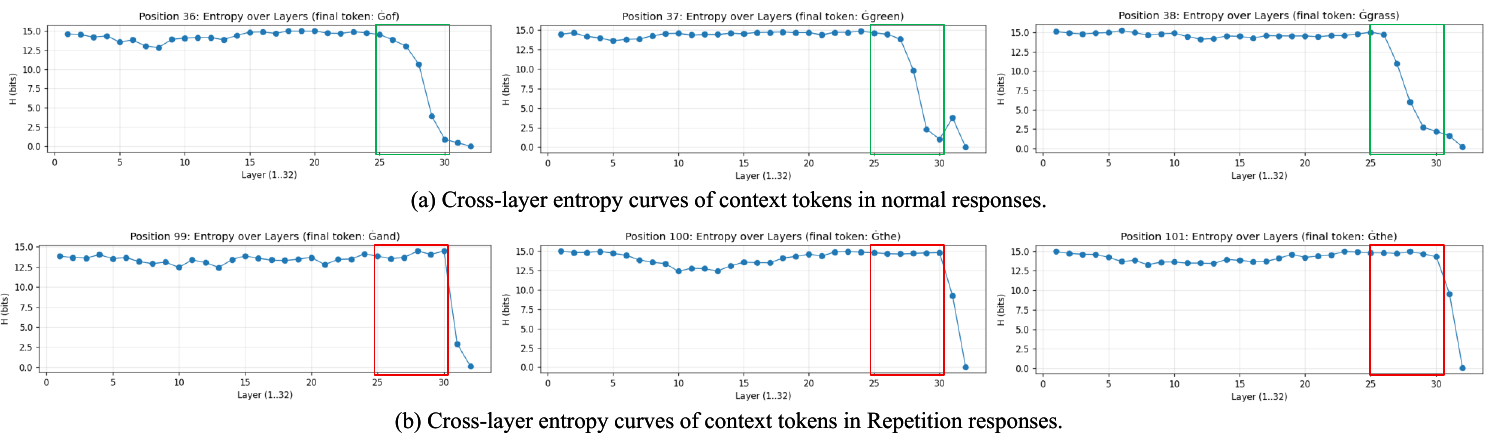}}
\caption{(a) and (b) correspond to context tokens from normal decoding and from decoding with repeated tokens, respectively. We define the set of context tokens as the target tokens together with their two nearest neighboring tokens in relative position. Context tokens with repetition tend to exhibit high entropy in deeper layers.}
\label{duibi}
\end{figure}

\subsection{Information Flow Analysis}\label{333}


Motivated by the need to understand the underlying cause of the Repeat Curse, we analyze the information flow to contrast model behaviors with and without cache, and provide a mechanistic interpretation of the phenomenon. To this end, we visualize the model’s attention matrices to analyze the information flow among tokens. As shown in Figure~\ref{context}, we observe that context tokens consistently receive high attention throughout the decoding process. Furthermore, attention progressively converges toward the context tokens from shallow to deeper layers. This phenomenon is reminiscent of the ‘attention sink’ observed in autoregressive models~\cite{fastv, Wang2023LabelWA, Zhang2024FromRT}, where certain special tokens act as anchors that aggregate information and absorb attention. The information flow pattern in Figure~\ref{context} highlights the role of context tokens as anchors in dMLLMs.

\vspace{-0.4cm}
\begin{findingbox}
\textbf{Finding 1: In dMLLMs, context tokens serve as anchors that aggregate information across layers and guide the final prediction.}
\end{findingbox}

\begin{figure}[t]
\centerline{\includegraphics[width=13cm]{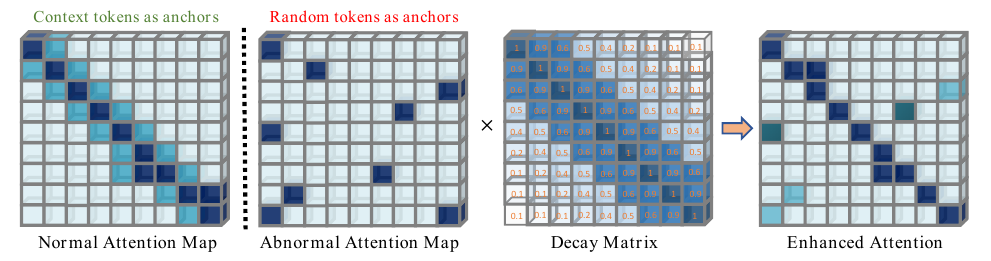}}
\caption{\textbf{Illustration of context tokens attention enhancement}. Example values are annotated on the decay matrix for clarity, while actual values are computed using \textcolor{red}{Equation~\ref{2} and ~\ref{1}.}}
\vspace{-0.4cm}
\label{ctae}
\end{figure}

Subsequently, we plot the cross-layer entropy curves of context tokens, as shown in Figure~\ref{duibi}. We find that the entropy of context tokens remains high in the shallow layers but gradually converges in the deeper layers, reflecting that as information accumulates layer by layer, the model’s predictions become increasingly stable and certain.

\vspace{-0.4cm}
\begin{findingbox}
\textbf{Finding 2: For context tokens under normal decoding, the information entropy gradually converges in the deeper layers.}
\end{findingbox}

Figure~\ref{Repeat}.b shows that after applying cache, the model exhibits a randomized attention distribution, disrupting the original information flow from context tokens and weakening output stability. Figure~\ref{duibi}.b further presents the cross-layer entropy curves of context tokens under repetition, revealing that repetition is accompanied by a failure of entropy convergence in the deeper layers.

\vspace{-0.4cm}
\begin{findingbox}
\textbf{Finding 3: Repetition is typically linked to disruptions in the information flow of context tokens and to the inability of their entropy to converge in deeper layers.}
\end{findingbox}

\section{Method}

Building on the observations in Section~\ref{moti}, we propose CoTA, a training-free method to mitigate the Repeat Curse, which has two key components: (1) Context Tokens Attention Enhancement, which preserves the intrinsic information flow pattern of context tokens, and (2) Context Tokens Entropy-Guided Voting, which prevents outputs driven by uncertain context tokens during decoding.

\subsection{Context Tokens Attention Enhancement (CATE)}

As analyzed in Section~\ref{moti}, context tokens in dMLLMs typically serve as anchors to aggregate information and absorb attention, but the use of cache disrupts this information flow pattern. We conjecture that tokens with shorter relative distances exhibit stronger semantic correlations, and thus the model’s increased attention to context tokens facilitates local semantic coherence. Therefore, we propose Context Tokens Attention Enhancement (CTAE), a simple yet effective attention intervention. We formalized the complete algorithmic procedure in Algorithm~\ref{Algorithm1}. As illustrated in the figure~\ref{ctae}, the core idea is to introduce a decay term and apply an element-wise multiplication with the attention matrix. Given a query token $q_i$ at position $i$ and a key token $k_j$ at position $j$, the decay term $\mathcal{G}_{i,j}$ for each query–key pair is computed as follows:
\begin{equation}
\mathcal{G}_{i,j} = \gamma_{\min} + (1-\gamma_{\min}) g_{i,j}, \quad \gamma_{\min} \in (0,1],
\label{2}
\end{equation}
\begin{equation}
g_{i,j} = 
\exp\!\left(-\left(\tfrac{|i-j|}{\tau}\right)^{2}\right).
\label{1}
\end{equation}
Here, $g_{i,j}$ denotes the Gaussian decay term computed using the exponential function $\exp()$ and the relative positional distance $|i-j|$ between query and key tokens. $\tau$ denotes the temperature factor, which is fixed to 5 in the experiments. We then introduce a lower-bound constant $\gamma_{\min}$ for stabilization, yielding the final decay term $\mathcal{G}_{i,j}$.

Assume the attention between a query token $q_i$ and a key token $k_j$ is denoted as $Attn_{i,j}$. We enhance the attention to context tokens by applying $Attn_{i,j} * \mathcal{G}_{i,j}$, thereby preserving the native information flow pattern in dMLLMs.

\begin{figure*}[t]
\centering
\begin{minipage}[t]{0.48\textwidth}
\begin{algorithm}[H]
\caption{Context Tokens Attention Enhancement (CTAE)}
\label{Algorithm1}
\begin{algorithmic}[1]
\Require Attention weights $A \in \mathbb{R}^{L \times H \times T \times T}$, temperature $\tau>0$, lower bound $\gamma_{\min}\in(0,1]$
\Ensure Enhanced attention $\tilde{A} \in \mathbb{R}^{L \times H \times T \times T}$
\State $\tilde{A} \gets A$
\For{$\ell \gets 1$ to $L$} \Comment{layer index}
  \For{$h \gets 1$ to $H$} \Comment{head index}
    \State $D_{i,j} \gets |i-j| \quad \forall i,j \in \{1,\dots,T\}$
    \State $g_{i,j} \gets \exp\!\bigl(- (D_{i,j}/\tau)^2\bigr)$
    \State $\mathcal{G}_{i,j} \gets \gamma_{\min} + (1-\gamma_{\min}) \cdot g_{i,j}$
    \State $\tilde{A}_{\ell,h} \gets A_{\ell,h} \odot \mathcal{G}$
  \EndFor
\EndFor
\State \textbf{return} $\tilde{A}$
\end{algorithmic}
\end{algorithm}
\end{minipage}\hfill
\begin{minipage}[t]{0.48\textwidth}
\begin{algorithm}[H]
\caption{Context Tokens Entropy-Guided Voting (CTEV)}
\label{algorithm2}
\begin{algorithmic}[1]
\State \textbf{Input:} Deep-layer logits $z^{(l)}_{i,:}$, base confidences $c_{(i)}$, coefficient $\alpha$
\For{$i \gets 1$ \textbf{to} $M$}
  \State $E_{\text{sum}}(i) \gets 0$
  \For{$l \in \{26,\dots,30\}$}
    \State $p^{(l)}_v = \exp(z^{(l)}_{i,v}) / \sum_{u=1}^V \exp(z^{(l)}_{i,u})$
    \State $E_{\text{sum}}(i) \gets E_{\text{sum}}(i) - \frac{1}{\log V}\sum_{v=1}^V p^{(l)}_v \log p^{(l)}_v$
  \EndFor
\EndFor
\State Build $\mathcal{C}(t)=\{t\}\cup\{\text{two nearest tokens of }t\}$
\For{$i \in \mathcal{C}(t)$}
  \State $E_{\text{sum}}^{ctx}(i) \gets \sum_{j \in \mathcal{C}(i)} E_{\text{sum}}(j)$
  \State $\mathrm{Score}(i) \gets c_{(i)} + \alpha \cdot E_{\text{sum}}^{ctx}(i)$
\EndFor
\end{algorithmic}
\end{algorithm}
\end{minipage}
\vspace{-0.2cm}
\end{figure*}

\textbf{Synergistic Workflow.} CTAE applies a decay term to attention values based on relative distance, enhancing the attention to context tokens. CTEV introduces deep-layer context token entropy as a penalty to the voting scores. Together, they jointly alleviate the repeat curse.

\subsection{Context Tokens Entropy-Guided Voting (CTEV)}

As observed in Section~\ref{moti}, context tokens with repetition typically exhibit persistently high entropy in deeper layers, reflecting the model’s uncertainty about context tokens during decoding. Yet baseline dMLLMs rely solely on confidence scores to vote for candidate decoding tokens, ignoring this uncertainty. To address this, we propose Context Tokens Entropy-Guided Voting (CTEV), which incorporates the aggregated deep-layer entropy of context tokens as a penalty term on the confidence score to prevent decoding under uncertain context tokens. The complete algorithm is formalized in Algorithm~\ref{algorithm2}. First, we compute the entropy of each candidate token based on its softmax probability distribution $p_{v}$ as follows:
\begin{equation}
E = \frac{-\sum_{v=1}^{V} p_{v}\,\log p_{v}}{\log V}, p_{v} = \frac{\exp(z_{v})}{\sum_{u=1}^{V} \exp(z_{u})},
\end{equation}
where $z_{v}$ denotes the logits of the $v$-th word in the vocabulary, and $V$ is the vocabulary size. On this basis, the entropy of tokens in deeper layers \footnote{Here, we define layers 26–30 as the deep layers, and target tokens together with their two nearest tokens in relative position are defined as the context tokens.} is accumulated layer by layer as follows:
\begin{equation}
E_{\text{sum}} 
= \sum_{l=26}^{30} E^{(l)}
= \sum_{l=26}^{30} \frac{-\sum_{v=1}^{V} p^{(l)}_{v}\,\log p^{(l)}_{v}}{\log V}.
\end{equation}
Let $E_{\text{sum}}(i)$ denote the $E_{\text{sum}}$ of token $x_i$ at position $i$ in a sequence of length $M$. The accumulated deep-layer entropy of the context tokens, $E_{\text{sum}}^{ctx}(i)$, is then computed as follows:
\begin{equation}
E_{\text{sum}}^{ctx}(i) \;=\; \sum_{j \in \mathcal{C}(i)} E_{\text{sum}}(i),\qquad 
\mathcal{C}(i) = \{\, x_i \,\} \cup 
\Big\{\, x_j \;\big|\; 
j \in \underset{j \in [M]\setminus\{i\}}{\operatorname{arg\,min}_2}\, |j-i| 
\Big\}.
\end{equation}
Finally, the weighted $E_{\text{sum}}^{ctx}(i)$ with coefficient $\alpha$ is incorporated into the confidence score $c_{(i)}$ from Equation~\ref{ci}, resulting in the new voting score $\text{Score}(i)$ defined as follows:
\begin{equation}
\text{Score}(i) = c_{(i)} + \alpha E_{\text{sum}}^{ctx}(i).
\label{11}
\end{equation}


\section{Experiments}


\subsection{Experimental Setup Details}\label{setDetails}

\noindent
\textbf{Model.} Following the baseline dMLLM LLaDA-V~\cite{lladav}, the language tower adopts LLaDA-8B-Instruct~\cite{llada}, the vision tower employs siglip2-so400m-patch14-384~\cite{Tschannen2025SigLIP2M}, and the projector is a two-layer MLP.

\noindent
\textbf{Evaluation.} We evaluate the effectiveness of our method on eight multimodal benchmarks: DocVQA~\cite{Mathew2020DocVQAAD}, ChartQA~\cite{Masry2022ChartQAAB}, MMStar~\cite{mmstar}, MME~\cite{mme}, Seed~\cite{Li2023SEEDBenchBM}, LLaVA$^w$~\cite{llava1.5}, MathVista~\cite{Lu2023MathVistaEM}, and MMBench~\cite{MMBench}. The Appendix~\ref{appendix1} reports more details about settings.

\subsection{Results on “repeat curse” evaluation}

We randomly select 500 samples from COCO2014~\cite{Lin2014MicrosoftCC} for the caption VQA task. The generated captions are then evaluated for mitigating the “Repeat Curse.” The evaluation metrics include Adjacent Repetition Rate (ARR), Maximum Repetition Length (MRL), Average Repetition Length (ARL), and Sample Repetition Rate (SRR). Table~\ref{repeat} presents the quantitative results of our method in mitigating repetition, evaluated under both long-text and short-text response settings. Experimental results are obtained by aggregating all samples and taking the average. The baseline model shows only minimal repetition, with the maximum repetition length being 2. In contrast, applying cache causes a rapid degradation of the outputs. Moreover, longer responses are more susceptible to repetition; for instance, with an output length of 512, ARR rises by 14.1 and SRR by 75.4. Our method demonstrates effective repetition mitigation, yielding ARR improvements of 13.1 and 6.1 under the long-response and short-response settings, respectively. We further perform ablation studies on two key components of our method, CTEV and CTAE. The results indicate that each component alone can also effectively mitigate repetition, highlighting the flexibility and complementarity of the method.


\begin{table}[t]
\centering
\setlength{\tabcolsep}{5pt}
\renewcommand{\arraystretch}{1.2}
\resizebox{\textwidth}{!}{%
\begin{tabular}{lcccccccc}
\toprule
Method & \multicolumn{4}{c}{512} & \multicolumn{4}{c}{64} \\
\cmidrule(lr){2-5} \cmidrule(lr){6-9}
& ARR$\downarrow$ & MRL$\downarrow$ & ARL$\downarrow$ & SRR$\downarrow$ 
& ARR$\downarrow$ & MRL$\downarrow$ & ARL$\downarrow$ & SRR$\downarrow$ \\
\midrule
LLaDA-V          & 0.2 & 2.0 & 2.0 & 6.9 & 0.1 & 2.0 & 2.0 & 3.3 \\
+ dllm-cache & 14.3 & 11.0 & 3.8 & 82.3 & 7.1 & 6.1 & 2.7 & 65.6 \\
+ dllm-cache + CTEV        & 3.2 & 2.2 & 2.0 & 10.6 & 2.5 & 2.3 & 2.1 & 5.6 \\
+ dllm-cache + CTAE        & 2.9 & 1.9 & 1.4 & 8.0 & 1.8 & 1.6 & 1.4 & 4.6 \\
+ dllm-cache + CTEV +CTAE  & \textbf{1.2} & \textbf{1.3} & \textbf{1.2} & \textbf{6.3} 
                           & \textbf{1.0} & \textbf{1.1} & \textbf{1.1} & \textbf{3.0} \\
\bottomrule
\end{tabular}%
}
\caption{\textbf{“Repeat Curse” Evaluation Results.} 512 and 64 denote the maximum generation length.}
\vspace{-0.3cm}
\label{repeat}
\end{table}

\begin{table*}[t]
\centering
\resizebox{\textwidth}{!}{
\begin{tabular}{lcc|cccccc}
\toprule
Model & Type & LLM Tower & DocVQA  & ChartQA & MMStar & MME$^p$  & Seed$^I$ & MMBench  \\
\midrule
LLaVA1.5 & AR & Vicuna-7B & -  & - & - & 1510 & 66.1 & 64.3  \\
Qwen2-VL & AR & Qwen2-7B & -  & 83.0 & 60.7 & - & - & -  \\
DeepSeek-VL & AR & DeepSeek-7B  & - & - & - & - & 70.4 & 73.2  \\
LLaVA-OV & AR & Qwen2-7B & -  & 80.0 & 61.7 & 1580 & 75.4 & 80.8  \\
MetaMorph & AR+Diff. & LLaMA3.1-8B  & - & 37.1 & - & - & 71.8 & 75.2  \\
JanusFlow & AR+Diff. & DeepSeek-1.3B  & - & 64.6 & - & 1333 & 70.5 & 74.9  \\
LLaMA3-V & Diff. & LLaMA3-8B & 86.2  & 77.8 & 56.5 & 1581 & 76.6 & 79.8  \\
\midrule
LLaDA-V & Diff. & LLaDA-8B & 83.9  & 78.3 & 60.1 & 1507 & 74.8 & 82.9  \\
LLaDA-V w/dllm-cache & Diff. & LLaDA-8B &  82.1 & 78.1 & 58.3 & 1410 & 72.1& 83.0  \\

 \rowcolor{blue!5}
Ours& Diff. & LLaDA-8B & 84.1  & 78.4 & 59.3 & 1523 & 73.9 & 83.1  \\
\bottomrule
\end{tabular}}
\caption{Benchmark results of different MLLMs on multiple multimodal evaluation datasets.}
\label{mllms}
\vspace{-0.25cm}
\end{table*}

\subsection{Comparison with Other MLLMs}

Table~\ref{mllms} presents a performance comparison with other paradigm MLLMs. Our goal is not absolute performance maximization, but rather to mitigate the output degradation of baseline models when using cache. Results across six benchmarks demonstrate the effectiveness of our method.

\subsection{Generality Study} 

We present the generalization evaluation of our method in Table~\ref{Generality}. Our approach consistently lowers response repetition across both open-domain natural and mathematical tasks, highlighting its generalizability. While caching reduces computational cost, it inevitably causes performance degradation, which our method successfully alleviates. Concretely, it yields gains of +6.7 in Score and +5.9 in ARR on MathVerse, and +1.8 in ACC and +5.6 in ARR on MathVista.

Our method can be flexibly applied as a complement to caching. We further evaluate its computational overhead, with experiments on LLaVAW demonstrating that it incurs only a 2.8 TPS reduction and a 1.9 FLOPs increase.


\begin{table}[t]
\centering
\setlength{\tabcolsep}{5pt}
\renewcommand{\arraystretch}{1.2}
\resizebox{\textwidth}{!}{
\begin{tabular}{lcccccccc}
\toprule
Method & \multicolumn{4}{c}{LLaVA$^{w}$} & \multicolumn{4}{c}{MathVista} \\
\cmidrule(lr){2-5} \cmidrule(lr){6-9}
& Score$\uparrow$  &  ARR$\downarrow$&TPS$\uparrow$& FLOPs $\downarrow$ & Score $\uparrow$ &  ARR $\downarrow$ &TPS $\uparrow$ & FLOPs $\downarrow$\\
\midrule
LLaDA-V & 70.1 & 1.3&  7.3 & 16.1& 59.7  & 0 &8.3& 14.1\\
 \rowcolor{blue!5}
Ours & 70.5 & 1.1&  5.1 & 17.8& 59.8  & 0 &7.5& 15.3\\
LLaDA-V w/dllm-cache & 63.2 & 7.3&  23.1 & 3.2& 54.9  & 6.9 &31.0& 3.9\\
 \rowcolor{blue!5}
Ours w/dllm-cache & 69.9 & 1.4& 20.3 & 5.1& 59.7  & 1.3 &28.7& 5.0\\
\bottomrule
\end{tabular}
}
\caption{\textbf{Generalization evaluation results of the method}. TPS stands for Tokens Per Second, and FLOPs stands for Floating Point Operations.}
\label{Generality}
\end{table}

\subsection{Ablation Study}\label{canshu}

We conduct an ablation study on the hyperparameters involved in our method, including the weighting factor $\alpha$ in Equation~\ref{11}, the minimum gain $\gamma_{\min}$ in Equation~\ref{2}, and the number of context tokens. We report the results in Table~\ref{hyperparameters}: the best hyperparameter configuration is $\alpha = 0.75$, $\gamma_{\min} = 0.5$, and the number of context tokens set to 3, achieving an ARR of 1.2\% and an ACC of 23.1. When the length of context tokens is set to 1, only the current target token is considered while surrounding tokens are ignored. We find that using longer context tokens (e.g., 5) does not lead to better results.


\begin{table*}[h]
\centering
\resizebox{\textwidth}{!}{%
\begin{tabular}{ccc}
\begin{subtable}[t]{0.32\textwidth}
\centering
\begin{tabular}{c c c c}
\toprule
$\gamma_{\min}$ & $\alpha$ & ARR$\downarrow$ & ACC$\uparrow$ \\
\midrule
\multirow{3}{*}{1} & 0.25 & 2.8 & 21.4 \\
                   & 0.5 & 2.2 & 21.9 \\
                   & 0.75 & 2.3 & 22.1 \\
\midrule
\multirow{3}{*}{0.5} & 0.25 & 2.9 & 20.0 \\
                     & 0.5 & 2.5 & 20.2 \\
                     & 0.75 & 2.1 & 20.1 \\
\bottomrule
\end{tabular}
\caption{Context tokens number = 1}
\end{subtable}
&
\begin{subtable}[t]{0.32\textwidth}
\centering
\begin{tabular}{c c c c}
\toprule
$\gamma_{\min}$ & $\alpha$ & ARR$\downarrow$ & ACC$\uparrow$ \\
\midrule
\multirow{3}{*}{1} & 0.25 & 2.1 & 22.1 \\
                   & 0.5 & 2.0 & 22.5 \\
                   & 0.75 & 1.7 & 22.3 \\
\midrule
\multirow{3}{*}{0.5} & 0.25 & 1.6 & 22.2 \\
                     & 0.5 & 1.4 & 22.0 \\
                     & 0.75 & 1.2 & 23.1 \\
\bottomrule
\end{tabular}
\caption{Context tokens number = 3}
\end{subtable}
&
\begin{subtable}[t]{0.32\textwidth}
\centering
\begin{tabular}{c c c c}
\toprule
$\gamma_{\min}$ & $\alpha$ & ARR$\downarrow$ & ACC$\uparrow$ \\
\midrule
\multirow{3}{*}{1} & 0.25 & 3.5 & 19.6 \\
                   & 0.5 & 3.8 & 19.2 \\
                   & 0.75 & 3.1 & 19.2 \\
\midrule
\multirow{3}{*}{0.5} & 0.25 & 3.9 & 19.0 \\
                     & 0.5 & 3.4 & 18.2 \\
                     & 0.75 & 3.3 & 19.1 \\
\bottomrule
\end{tabular}
\caption{Context tokens number = 5}
\end{subtable}
\end{tabular}%
}
\caption{Ablation results on hyperparameters evaluated on the MathVerse benchmark.}
\label{hyperparameters}
\end{table*}

\section{Conclusion and Limitations}

This paper investigates the phenomenon of repeated text generation in dMLLMs when using cache, which we term the “Repeat Curse.” Through information flow analysis, we reveal that in baseline dMLLMs, context tokens act as anchors to aggregate information and guide predictions. Applying cache disrupts this pattern, leading to repetition. Moreover, repetitive context tokens exhibit persistently high entropy in deeper layers. Building on these insights, we propose CoTA, a plug-and-play approach to mitigate the Repeat Curse. Extensive experiments demonstrate the effectiveness of the CoTA design. Due to the limited research on baseline dMLLMs, CoTA has not yet been validated for generalizability across more open-source dMLLMs and base models of different scales. Moreover, as cache methods for dMLLMs are still scarce, CoTA cannot be tested on a wider range of cache approaches. Future work will address these limitations.


\bibliography{iclr2026_conference}
\bibliographystyle{iclr2026_conference}

\appendix
\clearpage

\section{\textcolor{red}{Analysis of Cache Mechanism and Repetition Curse}}\label{rebuttal1}
\textcolor{red}{We conduct a series of ablation studies on LLaDA-V to analyze how different components of the cache mechanism influence repetition curse.}

\begin{table*}[t]

\label{tab:ablation_full}

\begin{minipage}{0.48\linewidth}
\centering
\subcaption{\textcolor{red}{Effect of prompt token recompute interval.}}
\label{tab:prompt_interval}
\begin{tabular}{lcccc}
\toprule
\textbf{Interval} & 1 & 5 & 15 & 25 \\
\midrule
\textbf{SRR}$\downarrow$ & 0 & 0 & 0 & 0 \\
\bottomrule
\end{tabular}
\end{minipage}\hfill
\begin{minipage}{0.48\linewidth}
\centering
\subcaption{\textcolor{red}{Effect of output token recompute interval.}}
\label{tab:output_interval}
\begin{tabular}{lcccc}
\toprule
\textbf{Interval} & 1 & 3 & 5 & 7 \\
\midrule
\textbf{SRR}$\downarrow$ & 0 & 79.9 & 87.4 & 89.7 \\
\bottomrule
\end{tabular}
\end{minipage}

\vspace{0.7em}

\begin{minipage}{0.48\linewidth}
\centering
\subcaption{\textcolor{red}{Effect of similarity thresholds.}}
\label{tab:sim_threshold}
\begin{tabular}{lccccc}
\toprule
\textbf{Threshold} & 0 & 0.25 & 0.5 & 0.75 & 1 \\
\midrule
\textbf{SRR} $\downarrow$& 89.7 & 75.0 & 69.8 & 29.7 & 0 \\
\bottomrule
\end{tabular}
\end{minipage}\hfill
\begin{minipage}{0.48\linewidth}
\centering
\subcaption{\textcolor{red}{Comparison of reuse policies.}}
\label{tab:reuse_policy}
\begin{tabular}{lcc}
\toprule
\textbf{Policy} & prefix KV cache & dllms-cache \\
\midrule
\textbf{SRR}$\downarrow$ & 0 & 75.0 \\
\bottomrule
\end{tabular}
\end{minipage}
\centering
\caption{\textcolor{red}{Ablation study on cache mechanism design components and their effects on repetition.}}
\end{table*}

\textcolor{red}{(a) We begin by examining the prompt token recomputation interval, while fixing the output token recomputation interval to 1 and setting the similarity threshold to 0. As shown in Table \ref{tab:prompt_interval}, periodic caching of prompt tokens has negligible impact on repetition.}

\textcolor{red}{(b) We then fix the prompt token recomputation interval to 25 and set the similarity threshold to 0, varying only the output token recomputation interval. As reported in Table \ref{tab:output_interval}, we observe that periodic caching of output tokens induces repetition, and longer recomputation intervals lead to higher repetition rates.}

\textcolor{red}{(c) In the dLLM-Cache framework, a subset of output tokens is adaptively recomputed at each step based on a similarity threshold. To analyze this further, we fix the prompt token recomputation interval to 25 and the output token recomputation interval to 7, while varying the similarity threshold. As shown in Table \ref{tab:sim_threshold}, the threshold plays a critical role: lower thresholds correspond to higher repetition rates.}

\textcolor{red}{(d) Finally, we compare dLLM-Cache with prefix KV cache~\cite{kwon2023efficient, lavida} to analyze the effect of different reuse policies. We fix the prompt token recomputation interval to 25, the output token recomputation interval to 7, and the similarity threshold to 0.25. As shown in Table \ref{tab:reuse_policy}, prefix KV cache, which only reuses cached states for prefix tokens, does not trigger repetition.}

\textcolor{red}{These four experiments reveal that the prompt recomputation interval has little effect on repetition behavior, while the output token recomputation interval, similarity threshold, and caching policy exert a substantial influence. In general, the fewer output tokens that are recomputed at each decoding step, the more likely the model is to fall into repetition.}

\textcolor{red}{We hypothesize that this is because the token states within the prompt sequence remain relatively static, so caching them has minimal impact on future predictions. In contrast, output tokens evolve across decoding steps. Caching these dynamic tokens forces the model to rely on outdated states, reducing uncertainty during prediction—ultimately contributing to the non-convergence of context-token entropy in deeper layers.}

\begin{table}[t]
\label{tab:repetition_all}
\begin{subtable}[t]{0.48\linewidth}
\centering
\begin{tabular}{lcc}
\toprule
Method & ARR$\downarrow$ & SRR$\downarrow$ \\
\midrule
MMaDA & 0.7 & 6.0 \\
+dllm-cache & 4.4 & 55.0 \\
+dllm-cache+CTAE & 2.4 & 29.0 \\
+dllm-cache+CTEV & 0.8 & 35.0 \\
+dllm-cache+CoTA & 0.6 & 30.0 \\
\bottomrule
\end{tabular}
\caption{\textcolor{red}{Repetition curse across different dMLLMs.}}
\label{tab:mmada_sub}
\end{subtable}
\hfill
\begin{subtable}[t]{0.48\linewidth}
\centering
\begin{tabular}{lcc}
\toprule
Method & ARR$\downarrow$ & SRR$\downarrow$ \\
\midrule
LLaDA-V & 0 & 0 \\
+dllm-cache & 9.3 & 75.0 \\
+prefix KV cache & 0 & 0 \\
LaViDa & 0 & 0 \\
+prefix KV cache & 0 & 0 \\
\bottomrule
\end{tabular}
\caption{\textcolor{red}{Repetition curse under different cache methods.}}
\label{tab:prefix_sub}
\end{subtable}
\centering
\caption{\textcolor{red}{Generalization experiments.}}
\end{table}

\section{\textcolor{red}{More Results}}\label{rebuttal2}

\subsection{\textcolor{red}{Repetition Curse Across Different dMLLMs}}\label{rebuttal2.1}

\textcolor{red}{We present the repetition analysis of applying the cache mechanism to MMaDA in Table \ref{tab:mmada_sub}. Notably, we find that MMaDA also suffers from the repetition curse when dLLM-Cache is enabled. Specifically, the adjacent repetition rate increases by +3.7\%, and the sample repetition rate increases by +49\%. These findings highlight that the repetition curse is not limited to a single architecture, but is a pervasive issue across dMLLMs.}

\begin{figure}[t]
\centerline{\includegraphics[width=13cm]{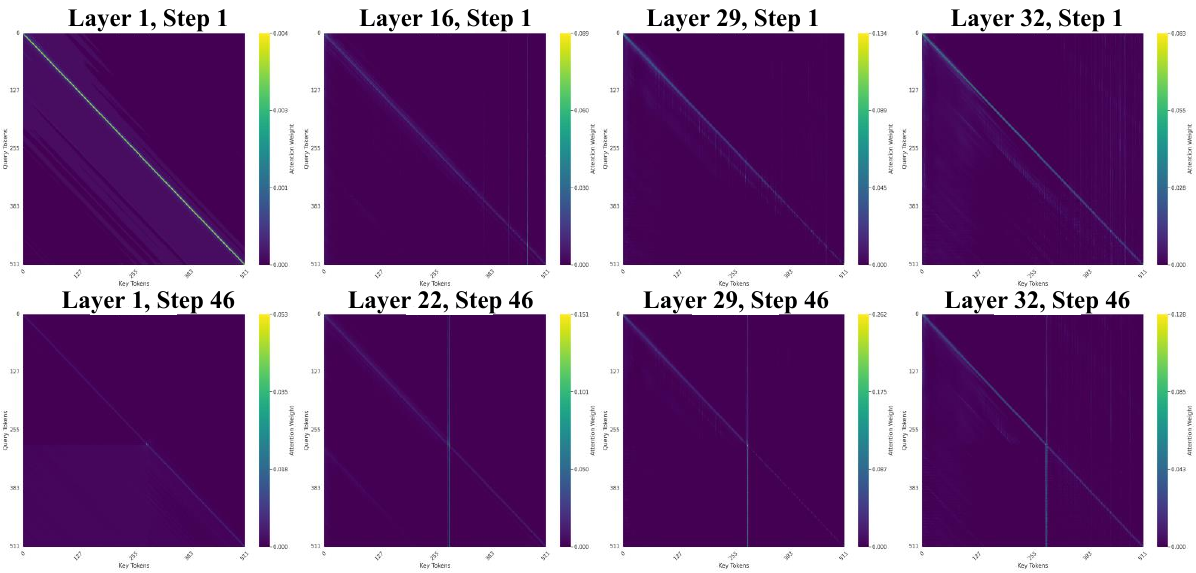}}
\caption{\textcolor{red}{Information flow visualization for MMaDA-8B without using cache.}}
\label{mmada-acontext}
\end{figure}

\begin{figure}[t]
\centerline{\includegraphics[width=13cm]{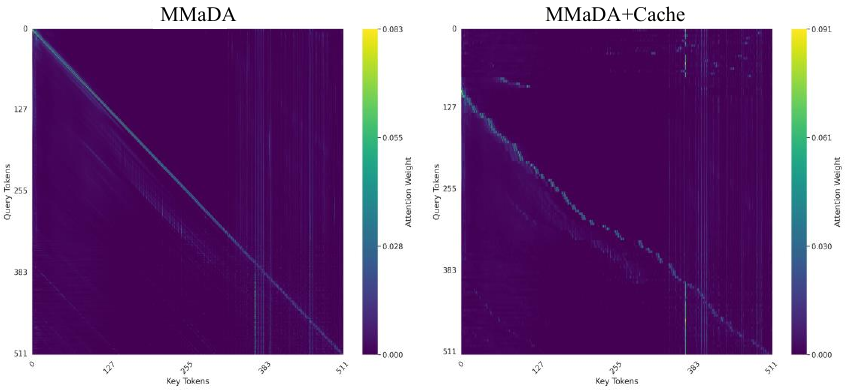}}
\caption{\textcolor{red}{Comparison of MMaDA's information flow patterns with and without cache.}}
\label{mmada-cache}
\end{figure}

\subsection{\textcolor{red}{Information Flow Across Different dMLLMs}}\label{rebuttal2.2}

\textcolor{red}{As illustrated in Figure \ref{mmada-acontext}, we visualize the attention maps of MMaDA to analyze the flow of information between tokens. We observe a similar pattern to LLaDA-V, where context tokens function as anchors, continually attracting attention across layers. Interestingly, we also detect vertical bands of concentrated attention in Layer 22 at Step 46, resembling the “attention sink”~\cite{OPERA} phenomenon commonly seen in autoregressive MLLMs. However, after introducing the cache, the original attention pattern of MMaDA is also disrupted (as shown in Figure \ref{mmada-cache}).}

\subsection{\textcolor{red}{Generalization Experiments of CoTA}}\label{rebuttal2.3}

\textcolor{red}{We apply CoTA to the MMaDA model, which exhibits repetition after enabling cache, and present the results in Table \ref{tab:mmada_sub}. The results show that CoTA and its components (CTAE and CTEV) effectively suppress repetitive text generation in MMaDA as well, demonstrating the generalizability of our approach.}

\subsection{\textcolor{red}{Repetition Curse Under Different Cache Methods}}\label{rebuttal2.4}

\textcolor{red}{As shown in Table \ref{tab:prefix_sub}, we further assess the repetition behavior when applying the prefix-KV cache method ~\cite{kwon2023efficient} to both LLaDA-V and LaViDa~\cite{lavida}. Notably, the prefix-KV cache does not trigger repetition in either model. We speculate that this is because prefix-KV cache only reuses the cached states of prefix tokens (prompt and image tokens), without affecting suffix tokens (output text tokens). This finding is consistent with the conclusions presented in Section \ref{rebuttal1}.}

\begin{figure*}[t]
    \centering

    \begin{subfigure}{0.47\textwidth}
        \centering
        \includegraphics[width=\textwidth]{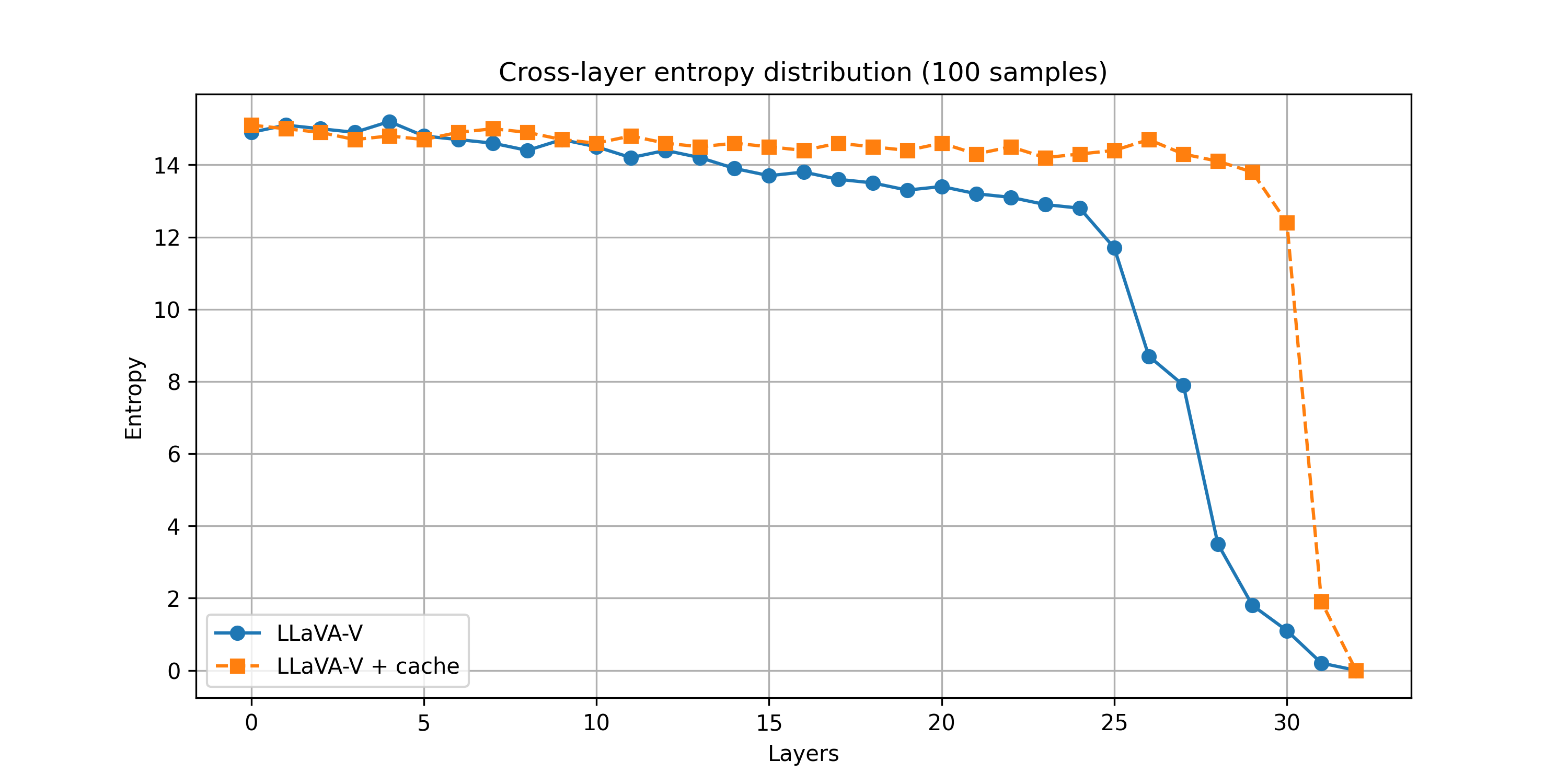}
        \caption{\textcolor{red}{Cross-layer entropy distribution across 100 samples.}}
        \label{entropy_plot}
    \end{subfigure}
    \hfill
    \begin{subfigure}{0.47\textwidth}
        \centering
        \includegraphics[width=\textwidth]{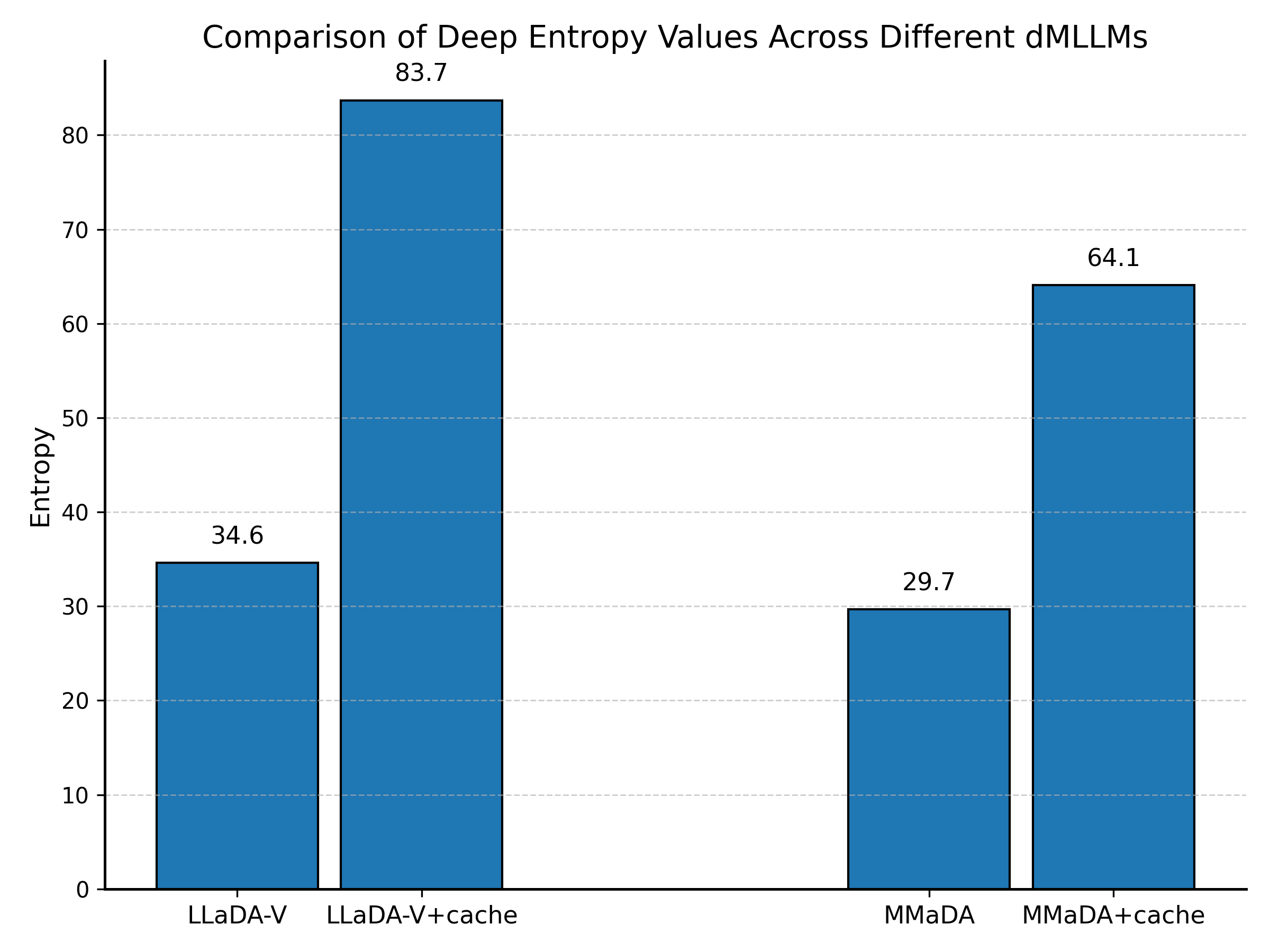}
        \caption{\textcolor{red}{Comparison of total deep-layer entropy across different dMLLMs.}}
        \label{entropy_model}
    \end{subfigure}

    \caption{\textcolor{red}{Deep-layer entropy analysis experiment.}}
    \label{fig:mmada_combined}
\end{figure*}

\section{\textcolor{red}{More experimental details of CTEV}}\label{rebuttal3}

\textcolor{red}{In CTEV, a central design choice is the selection of layers 26–30 for entropy estimation. This choice is informed by both multi-sample statistics and detailed single-sample analyses. Concretely, we compute per-layer entropy over 100 samples, and then aggregate and average the entropy values for each layer across samples. The resulting entropy distribution across layers is shown in Figure \ref{entropy_plot}. A comparison of the entropy trajectories before and after enabling the cache reveals a clear pattern: cache usage induces pronounced entropy deviations specifically in layers 26–30. As discussed in Section \ref{333}, these layers consistently exhibit the strongest sensitivity to cache-induced shifts, which directly motivates their adoption in CTEV.} 

\textcolor{red}{To further validate this choice, we examine the cumulative entropy over layers 26–30 across different dMLLM architectures. As shown in Figure \ref{entropy_model}, all models display a consistent trend: caching systematically disrupts entropy convergence in deep layers. This cross-model agreement provides strong empirical support for focusing on layers 26–30 within the CTEV framework.}

\begin{table}[h]
\centering
\begin{tabular}{lcccc}
\toprule
\textbf{Methods} & \textbf{16} & \textbf{32} & \textbf{64} & \textbf{128} \\
\midrule
LLaDA-V          & 2\%   & 1\%   & 1\%   & 1\%   \\
LLaDA-V+cache    & 88\% & 85\% & 79\%  & 75\%  \\
LLaDA-V+CoTA     & 17\%  & 14\%  & 10\%   & 8\%   \\
\bottomrule
\end{tabular}
\caption{\textcolor{red}{Repeat curse comparison under different block lengths. The evaluation metric is the sample repetition rate.}}
\label{tab:block_length}
\end{table}

\textcolor{red}{We further analyze the sensitivity of CTEV through additional experiments. As shown in Table \ref{tab:block_length}, we evaluate the CoTA performance under different block lengths during decoding. We observe that our method consistently reduces the repetition rate across all tested block configurations, demonstrating its robustness to different window sizes.}

\begin{table}[h]
\centering
\begin{tabular}{lcccc}
\toprule
\multirow{2}{*}{\textbf{Layer range}} 
& \multicolumn{2}{c}{\textbf{512}} 
& \multicolumn{2}{c}{\textbf{128}} \\
\cmidrule(lr){2-3} \cmidrule(lr){4-5}
& \textbf{ARR$\downarrow$} & \textbf{SRR$\downarrow$} 
& \textbf{ARR$\downarrow$} & \textbf{SRR$\downarrow$} \\
\midrule
LLaDA-V+cache & 13.9\% & 85\% & 9.3\% & 75\% \\
1--10         & 12.1\%   & 81\% & 8.4\% & 73\% \\
11--20        & 12.4\%   & 83\% & 9.1\% & 74\% \\
21--25        & 11.3\%   & 73\% & 7.9\% & 65\% \\
25--30        & 5.6\%   & 23\% & 4.3\% & 19\% \\
26--30        & 2.9\%  & 10\% & 1.8\% & 8\%  \\
26--31        & 3.2\%  & 12\% & 2.1\% & 10\% \\
26--32        & 3.3\%  & 12\% & 2.1\% & 9\% \\
\bottomrule
\end{tabular}
\caption{\textcolor{red}{Performance comparison under different layer ranges.}}
\label{tab:layer_block_comparison}
\end{table}

\textcolor{red}{We further present results across different layer ranges in Table \ref{tab:layer_block_comparison}. We find that computing the cumulative entropy using layers 26–30 yields the best performance. We speculate that this is because the entropy values in this depth range exhibit the largest separation between normal and abnormal modes, while incorporating additional layers introduces unnecessary noise, leading to diminished performance.}

\begin{figure}[t]
\centerline{\includegraphics[width=13cm]{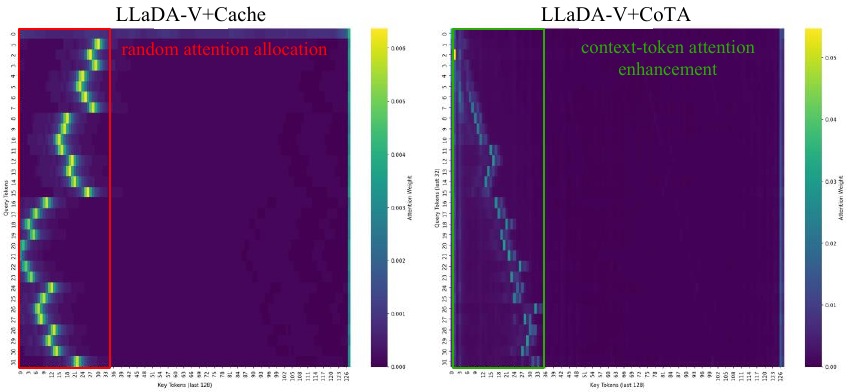}}
\caption{\textcolor{red}{Attention map visualizations of LLaDA-V+Cache and LLaDA-V+CoTA.}}
\label{ctaeduibi}
\end{figure}

\section{\textcolor{red}{Additional Analyses of CoTA}}

\textcolor{red}{As shown in Figure \ref{ctaeduibi}, we visualize the attention maps to analyze the changes in the model’s internal information-flow patterns. After applying Cache, the model exhibits a more random attention distribution, whereas CoTA restores the “context tokens as anchors” information-flow pattern by strengthening attention toward contextual tokens.}

\begin{table}[t]
\centering
\begin{tabular}{lccc}
\toprule
 & \multicolumn{3}{c}{LLaDA-V+cache} \\
\cmidrule(lr){2-4}
\textbf{Repeated word} & \textbf{the} & \textbf{of} & \textbf{a} \\
Repetition ratio   & 98\%         & 76\%        & 56\%        \\
\bottomrule
\end{tabular}
\caption{\textcolor{red}{Statistics of repeated words generated by LLaDA-V with cache.}}
\label{repeatedwords}
\end{table}

\section{\textcolor{red}{Linguistic Attribute Analysis of Repeated Tokens}}

\textcolor{red}{In this section, we analyze the linguistic attributes of repeated tokens by computing their frequency statistics over 200 samples that exhibit repetition. As shown in Table \ref{repeatedwords}, the top-3 most frequently repeated words are “the,” “of,” and “a.” We observe that repeated tokens are predominantly function words carrying low semantic content, which is also consistent with the case study presented in Figure \ref{motivation}. This indicates a common tendency of the model when generating uncertain or repetitive text.}

\begin{table*}[t]
\centering
\begin{subtable}{0.47\linewidth}
\centering
\begin{tabular}{lcc}
\toprule
\textbf{Methods} & \textbf{ACC}$\uparrow$ & \textbf{ARR}$\downarrow$ \\
\midrule
LLaDA-V+cache & 54.9 & 6.9 \\
n-gram (n=2)  & 54.0 & 2.4 \\
n-gram (n=3)  & 54.3 & 3.2 \\
CoTA          & 59.7 & 1.3 \\
\bottomrule
\end{tabular}
\caption{Comparison between CoTA and n-gram penalties.}
\label{tab:mathvista_sub}
\end{subtable}
\hfill
\begin{subtable}{0.47\linewidth}
\centering
\begin{tabular}{p{3.6cm}cc}
\toprule
\textbf{Methods} & \textbf{ARR}$\downarrow$ & \textbf{SRR}$\downarrow$ \\
\midrule
Decay Matrix & 1.8\% & 8\% \\
AliBI Bias Matrix & 3.9\% & 10\% \\
\bottomrule
\end{tabular}
\caption{Comparison of different attention biasing methods.}
\label{tab:ctae_ablation_sub}
\end{subtable}

\caption{\textcolor{red}{Experimental results of CoTA and related techniques.}}
\label{tab:combined_tables}
\end{table*}

\section{\textcolor{red}{Discussion of CoTA and Related Techniques}}

\textcolor{red}{N-gram penalties \cite{Holtzman2020The}, as a simple and widely used decoding-time technique, are commonly applied to control repetition. We further compare the performance of CoTA and n-gram penalties in Table \ref{tab:mathvista_sub}. We find that although n-gram penalties can reduce repetition, their effectiveness is inferior to that of CoTA. Moreover, applying n-gram penalties often leads to a degradation in model accuracy, likely due to forced token substitution, which disrupts semantic coherence and lowers output quality. In addition, we replace the decay matrix used in CTAE’s attention biasing design with an Alibi bias penalty matrix~\cite{Tang_2025_CVPR}to evaluate its effect. As shown in Table \ref{tab:ctae_ablation_sub}, the results confirm that using the Decay Matrix for attention intervention in CTAE is more effective.}

\begin{table}[t]
\centering
\begin{tabular}{lcc}
\hline
\textbf{Method} & \textbf{throughput (tok/s) $\uparrow$} & \textbf{latency (s/sample) $\downarrow$} \\
\hline
LLaDA-V               & 1.8  & 17.4 \\
+prefix KV cache      & 2.7  & 13.3 \\
+dllm-cache           & 4.9  & 6.1  \\
+dllm-cache+CTEV      & 4.1  & 7.2  \\
+dllm-cache+CTAE      & 4.6  & 6.4  \\
+dllm-cache+CoTA      & 3.8  & 7.7  \\
MMaDA                 & 0.5  & 32.1 \\
+dllm-cache           & 2.0  & 19.6 \\
+dllm-cache+CTEV      & 1.5  & 22.0 \\
+dllm-cache+CTAE      & 1.7  & 21.4 \\
+dllm-cache+CoTA      & 1.3  & 24.0 \\
\hline
\end{tabular}
\caption{\textcolor{red}{Throughput and latency comparison on DocVQA.}}
\label{tab:docvqa-efficiency}
\end{table}

\label{Performancelatency}
\section{\textcolor{red}{More Efficiency Analysis}}

\textcolor{red}{We further perform a comprehensive efficiency analysis on DocVQA, covering prefix-KV, dLLM-cache, CoTA, and its two components (CTAE and CTEV). As shown in Table \ref{Performancelatency}, integrating CoTA introduces only a modest and acceptable increase in latency, along with a slight reduction in throughput, yet it still delivers substantial improvements over the baseline. Moreover, to provide a more holistic assessment of efficiency, we additionally report latency and throughput results on both LLaDA-V and MMaDA.}

\section{\textcolor{red}{Related Work}}


\noindent
\textbf{Diffusion-based Multimodal Large Language Models (dMLLMs).} The latest diffusion-based large language models (dLLMs)~\cite{llada, llada1.5, dream2025, gong2025scaling} have been successfully scaled to 8B parameters, achieving performance comparable to state-of-the-art autoregressive large language models~\cite{llama, llama2, devlin2018bert, llama3, deepseek, qwen2.5, phi, vicuna}. Furthermore, by integrating visual instruction tuning and architectural extensions, dLLMs have recently been extended to dMLLMs~\cite{llada, llada1.5, dream2025, gong2025scaling}, demonstrating promising multimodal capabilities.

\noindent
\textbf{Diffusion Language Models.} Motivated by the remarkable success of diffusion models in image generation~\cite{ldm, SDXL, esser2024scaling}, recent studies have introduced the diffusion process into language modeling, including both continuous diffusion models~\cite{LovelaceKWSW23, li2022diffusionlm, zhang2025target, xue2024unifying, LinGSWFLDC23} and discrete diffusion models~\cite{campbell2022a, zheng2025masked, gat2024discrete, Ye2023DiffusionLM, Zheng2023ARD}. The latest masked diffusion models~\cite{llada, llada1.5, dream2025, gong2025scaling} have successfully scaled dLLMs to 8B parameters and demonstrated performance comparable to autoregressive large language models~\cite{ qwen2.5}.


\noindent
\textbf{Multimodal Large Language Models (MLLMs).} Building upon the unprecedented generative capabilities of large language models (LLMs)\cite{llama, llama2, devlin2018bert, llama3, deepseek, qwen2.5, phi, vicuna}, MLLMs have emerged as powerful systems that extend LLM architectures to process over multimodal inputs. Probabilistic modeling approaches for MLLMs generally fall into three paradigms: (i) autoregressive\cite{llava, llava1.5, Qwen2.5-VL, Qwen2-VL, internvl3, internvl, GPT4V, li2025llavaonevision}, (ii) autoregressive–diffusion hybrid~\cite{uniDiffuser, xie2025showo, zhou2025transfusion, JanusFlow, dimple}, and (iii) pure diffusion~\cite{swerdlow2025unidisc, li2025dualdiffusionunifiedimage}. Building on the recent breakthroughs in dLLMs, the latest dMLLMs~\cite{lladav, mmada, lavida} exploit the language modeling capabilities of dLLMs, coupled with effective training pipelines, to deliver performance on par with leading autoregressive and hybrid counterparts.

\noindent
\textbf{\textcolor{red}{Repeated Token Generation.}} \textcolor{red}{Early research on text repetition and degeneration in model predictions primarily focused on improving sampling strategies for language models \cite{Holtzman2020The}, such as nucleus sampling and top-k sampling. Several studies treat repetition control as a training objective and introduce various training-time strategies to mitigate repetitive generation \cite{Welleck2020Neural, lagutinmplicit, DITTO}. Other methods focus on incorporating repetition-aware penalties during the sampling stage \cite{zhu2023penalty}.}

\textcolor{red}{In recent years, research on repetitive text generation has expanded from conventional language generation models to modern LLMs. \cite{Frustratingly} estimates the likelihood of future repetition using an n-gram LM constructed from the generated prefix and imposes penalties based on this estimate. \cite{LZPenalty} introduces a repetition penalty from a compression-based perspective, while \cite{li2023repetition} highlights how repeated tokens in training corpora can exacerbate repetition during inference.}

\textcolor{red}{Some recent work further investigates the underlying mechanisms behind repetition. DUC \cite{Yao2025UnderstandingTR} employs Sparse Autoencoders (SAEs) to identify latent features that become highly activated when a model produces repeated tokens, referred to as repetition features. They attribute repetition to the activation of these features at specific layers and mitigate repetition by suppressing them. \cite{WangLL0S024} employs model-editing techniques to locate FFN neurons strongly associated with repetition, as well as neurons strongly related to the main task. By intersecting these neuron sets, they filter out neurons purely tied to repetition and perform targeted edits.}

\textcolor{red}{Distinct from the above approaches, our work examines repetition from an information-flow perspective. By comparing the information-flow patterns that emerge during repetitive generation with those observed under normal conditions, we identify the underlying mechanisms that lead to repetition. Our analysis shows that in dMLLMs, context tokens act as anchors that aggregate and propagate information. However, the introduction of cache disrupts this aggregation pattern and consequently triggers repetitive outputs. In addition, our layer-wise entropy analysis reveals that context tokens involved in local repetition exhibit a failure of entropy convergence in deeper layers. It is important to note that existing research on repetitive outputs in MLLMs and dMLLMs remains limited. Prior studies on MLLMs discuss repetition primarily in the context of hallucination and demonstrate that certain decoding strategies can reduce this effect \cite{tong2025mitigatinghallucinationmultimodalllms, OPERA}. Recent dMLLM work has reported the presence of repetition but has not conducted deeper investigation into its causes \cite{dimple}.}

\section{Evaluation configuration details of different datasets}\label{appendix1}

We report detailed information about the evaluation setup in Table \ref{tab:training_config}, including the maximum generation length, block length, decoding steps, and batch size for different benchmarks.

\begin{table*}[t]
\centering
\begin{tabular}{l|ccccc}
\hline
\textbf{Evaluation data} & \textbf{DocVQA} & \textbf{ChartQA} & \textbf{MMStar} & \textbf{MME} & \textbf{SEED}  \\
\hline
Max generation length & 32 & 16 & 2 & 2 & 2  \\
Block length & 32 & 16 & 2 & 2 & 2  \\
Decode steps & 16 & 16 & 2 & 2 & 2  \\
Batchsize & 1 & 1 & 1 & 1 & 1  \\
\hline
\end{tabular}
\caption{Evaluation configuration details of different datasets.}
\label{tab:training_config}
\end{table*}

\section{Maximum Repetition Length, Average Repetition Length, 95th-percentile Repetition Length, and Sample Repetition Rate}\label{appendix2}

Given an input sequence of length $M$, $\{\mathcal{T}_i\}_{i=1}^M$, we record the length of each consecutive identical segment $r_k$ as follows:
\begin{equation} 
r_k = \left|\{\, \mathcal{T}_i \;\mid\; \mathcal{T}_i = \mathcal{T}_{i+1} = \dots = \mathcal{T}_{i+r_k-1}\,\}\right|, \quad k = 1,2,\dots,K, 
\end{equation}
where $K$ denotes the total number of segments. Furthermore, we obtain all token repetition segments $rep_runs$ as follows:
\begin{equation} 
rep\_runs = \{\, r_k \;\mid\; r_k \geq 2,\; k=1,2,\dots,K \,\}. 
\end{equation}
Based on rep\_runs, the maximum repetition length (MRL), average repetition length (ARL), and 95th-percentile repetition length (95pRL) can be obtained from Equations~\ref{MRL}, ~\ref{ARL}, and ~\ref{95}, respectively.

\begin{equation} 
\text{MRL} = \max(\text{rep\_runs}).
\label{MRL}
\end{equation}

\begin{equation} 
\text{ARL} = \frac{1}{|\text{rep\_runs}|} \sum_{r \in \text{rep\_runs}} r.
\label{ARL}
\end{equation}

\begin{equation} 
\text{95pRL} = \operatorname{Quantile}_{0.95}(\text{rep\_runs}). 
\label{95}
\end{equation}

Let the total number of generated results be $N$, among which the number of results containing repetition is $N_{\text{dup}}$. The Sample Repetition Rate is defined as:
\begin{equation}
\text{SRR} = \frac{N_{\text{dup}}}{N}
\end{equation}

\section{Case Study}

As illustrated in the Figure~\ref{motivation}, incorporating dllm-cache into LLaDA-V leading to generated responses with excessive repetition of words (e.g., “the”) and punctuation marks (e.g., “,”). In contrast, our mitigation strategy substantially reduces such repetition. Notably, it also encourages the model to attend more effectively to the information contained in surrounding tokens, enabling the generation of image descriptions that are both more detailed and more coherent.

\begin{figure}[t]
\centerline{\includegraphics[width=13cm]{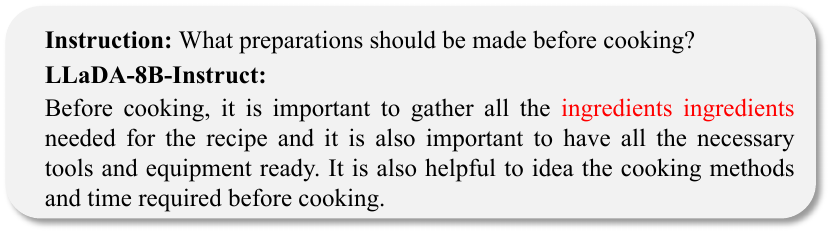}}
\caption{Repetition phenomena in dLLMs.}
\label{Repedllms}
\end{figure}

\section{Repetition Phenomena in dLLMs}

As illustrated in Figure~\ref{Repedllms}, we find that diffusion-based large language models likewise suffer from repetition in generated text. This finding motivates future research into the internal mechanisms of both dLLMs and dMLLMs. Furthermore, our study reveals that the repeated words are predominantly low-semantic terms (e.g., “the,” “is”), while the example in the Figure~\ref{Repedllms} suggests that repetitions in dLLMs may involve a broader variety of words.


\section{Declaration of the Use of Generative AI (Large Language Models)}

Generative AI tools, including Grammarly and ChatGPT, were used solely for grammar checking and language polishing. All technical content, experimental design, data analysis, and conclusions were generated and verified exclusively by the human authors. The use of AI tools does not affect the originality or authorship of this work.

\section{Ethics Statement}

This work seeks to uncover the underlying causes of the Repeat Curse in dMLLMs with cache from an information-flow perspective and to propose mitigation strategies that enhance the output performance of diffusion-based multimodal large language models (dMLLMs). The proposed methods, CoTA, is developed using only publicly available dMLLMs (LLaDA-V), caching approaches (dLLM-Cache), and benchmark datasets (e.g., MME, MMBench, SEED), without relying on any private, sensitive, or human-subject data. These techniques do not introduce biases beyond those inherent to the base models, and are designed to supplement rather than replace human supervision in critical applications. Nevertheless, while our methods can effectively mitigate response repetition, they cannot guarantee the elimination of errors or misleading outputs. Hence, caution is advised in practical deployments.

\section{Reproducibility Statement}
To ensure full reproducibility, we provide the following resources:
(1) Code: The complete implementation of CoTA, including Context Tokens Attention Enhancement (CTAE) and Context Tokens Entropy-Guided Voting (CTEV), will be publicly released on GitHub upon publication.
(2) Hyperparameters: For dLLM-Cache, all experiments fix the hyperparameters at $\alpha = 25\%$, $\mathcal{E}_p = 25$, and $\mathcal{E}_s = 7$. The parameter details for CoTA are described in Section~\ref{canshu}, while those for the evaluation setup are provided in Section~\ref{appendix1}.
(3) Evaluation: We use standard, publicly available benchmarks (e.g., MME, MMBench, SEED) along with their official evaluation scripts.
(4) Compute: Experiments are conducted on NVIDIA A800 80GB GPUs. Inference latency is reported using TPS and FLOPs (Table~\ref{Generality}).
(5) Models and Cache Methods: We evaluate open-source dMLLMs, specifically LLaDA-V (8B). The cache method is based on the official implementation of dLLM-Cache for LLaDA-V. No proprietary data or models are used in this work.

\end{document}

%% file: math_commands.tex
\usepackage{amsmath,amsfonts,bm}

\def\eqref#1{equation~\ref{#1}}

\def\1{\bm{1}}

\DeclareMathAlphabet{\mathsfit}{\encodingdefault}{\sfdefault}{m}{sl}
\SetMathAlphabet{\mathsfit}{bold}{\encodingdefault}{\sfdefault}{bx}{n}

